\documentclass[runningheads]{llncs}
\usepackage{graphicx}
\usepackage{authblk}

\begin{document}
\title{Embedding Task Knowledge into 3D Neural Networks via Self-supervised Learning}
% \email{**@******.***}
\author{Jiuwen Zhu\inst{1}\and
Yuexiang Li\inst{2} \and
Yifan Hu\inst{2} \and
S. Kevin Zhou\inst{1} }
\institute{Institute of Computing Technology, Chinese Academy of Sciences, Beijing, China\\ \email{zhoushaohua@ict.ac.cn}\\
\and Tencent Jarvis Lab, Shenzhen, China\\ \email{vicyxli@tencent.com}}


\maketitle              % typeset the header of the contribution

\begin{abstract}
  Deep learning highly relies on the amount of annotated data. However, annotating medical images is extremely laborious and expensive. To this end, self-supervised learning (SSL), as a potential solution for deficient annotated data, attracts increasing attentions from the community. However, SSL approaches often design a proxy task that is not necessarily related to target task. In this paper, we propose a novel SSL approach for 3D medical image classification, namely Task-related Contrastive Prediction Coding (TCPC), which embeds task knowledge into training 3D neural networks. The proposed TCPC first locates the initial candidate lesions via supervoxel estimation using simple linear iterative clustering. Then, we extract features from the sub-volume cropped around potential lesion areas, and construct a calibrated contrastive predictive coding scheme for self-supervised learning. Extensive experiments are conducted on public and private datasets. The experimental results demonstrate the effectiveness of embedding lesion-related prior-knowledge into neural networks for 3D medical image classification.
% 145 words 150-200

\end{abstract}
\section{Introduction}
The performance of deep learning networks strongly relies on the amount of annotated data. However, annotating 3D volumetric medical data requires experienced physicians to spend hours or days, which is laborious and expensive. Therefore, the amount of annotated 3D medical data is often insufficient for the training of 3D neural networks. To address the problem, researchers have attempted to loose the requirement of annotated data for neural networks by exploiting the rich information contained in unlabeled data \cite{Scaling2019}.

Self-supervised learning (SSL), as one of the potential solutions, gains increasing attentions from the community. Conventional self-supervised learning approaches start with a formulated proxy task to encourage neural networks to learn informative features from the raw data. Various proxy tasks have been proposed, such as grayscale image colorization \cite{larsson_colorization_2017}, images rotation \cite{gidaris2018image_rotations} and Jigsaw puzzles \cite{noroozi2016jigsaw_puzzles,Arbitrary_Jigsaw_Puzzles}. However, those proxy tasks primarily focus on the pre-training of 2D neural networks. Compared to 2D networks, 3D neural networks have shown their superiority for 3D volumetric data. Till now, there are few 3D self-supervised learning approaches. Rubik’s cube proposed by Zhuang et al.~\cite{Zhuang_2019_MICCAI} is one of the earliest attempts in the area. It formulates the volumetric data as a Rubik's cube and enforces 3D neural networks to learn anatomical information by playing a disarrangement-and-restoration game. Those existing 3D-based SSL approaches prefer to learn the feature representation more related to the common prior-knowledge, e.g., anatomical structure of organs. The information often used for clinical diagnosis, such as position and texture of potential lesion areas, is not explicitly embedded during the 3D network pre-training.

Contrastive predictive coding (CPC) \cite{denoord2018representation,henaff2019dataefficient}, which learns the feature representation of spatial information, is a recent SSL approach. The underlying principle of CPC is to use a feature set, extracted from patches cropped from part (upper/left) of 2D image, to predict the encoded features of the rest part (lower/right). The spatial information of image content is embedded during the predictive coding. However, due to the gap between 2D natural images and 3D medical volumes, the conventional CPC cannot be directly applied to 3D volumes.

In this paper, we aim to bridge this gap by proposing Task-related CPC (TCPC), which embeds the lesion information into the pre-training process. It initially locates the potential lesion areas using the simple linear iterative clustering (SLIC) \cite{achanta2012slic}, and then calibrates CPC to learn 3D feature representation from the sub-volumes containing the candidate lesion areas. To validate the effectiveness of our TCPC, we conduct experiments on public and private datasets. The experimental results demonstrate the benefits generated by embedding the clinical prior-knowledge (e.g., the texture and position of lesion area) to 3D neural networks for some specific tasks, e.g., brain hemorrhage classification.

\begin{figure}[t]
  \begin{center}
    \includegraphics[width=0.85\columnwidth]{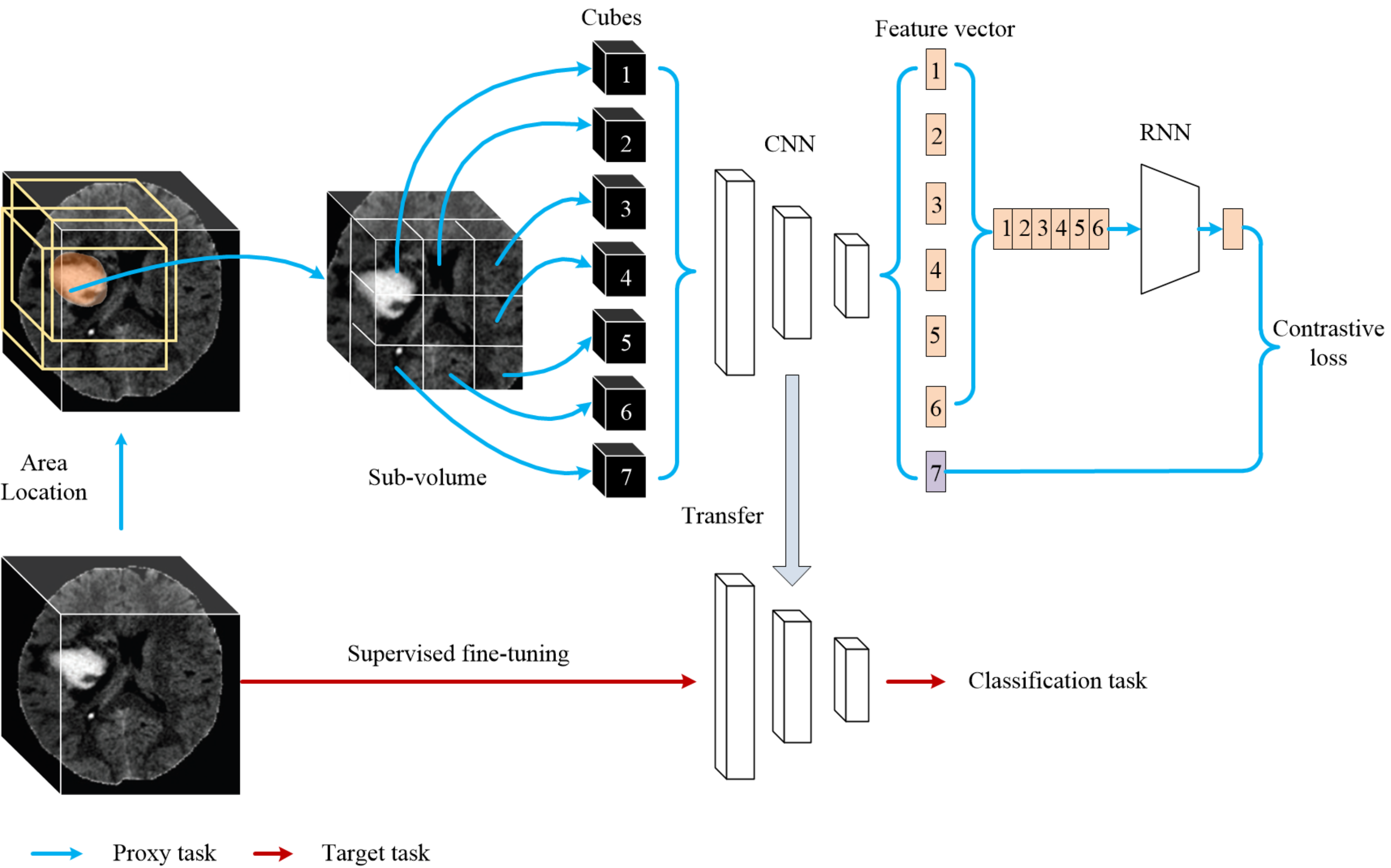}
    \caption{The proposed self-supervised learning framework. It consists of two stages: unsupervised feature learning and supervised fine-tuning.} 
    \label{framework}
  \end{center}
\end{figure}

\section{Task-related Constrastive Predictive Coding}
In this section, we introduce the proposed TCPC in details. As shown in Fig.~\ref{framework}, the pipeline of our self-supervised learning framework consists of two steps: 1) pre-train a 3D convolutional neural network (CNN) on our TCPC proxy task using unlabeled datasets; 2) fine-tune the pre-trained network for the specific target task with manual annotations.

\subsection{Supervoxel generation and subvolume cropping} 
For a raw 3D medical volume, the voxels are first normalized to [-1, 1] by using the minimum and maximum intensity. Then, the simple linear iterative clustering (SLIC)~\cite{achanta2012slic} method is utilized to generate the supervoxels, which are small clusters of pixels that share similar properties, e.g., grayscale intensity. The potential lesion areas can thereby be roughly detected and these areas are more meaningful for our target classification task than say pure background areas. Finally, we crop the sub-volumes centered on the region of each supervoxel for our calibrated CPC to unsupervisedly extract feature representations.

\subsection{Calibrated 3D contrastive predictive coding}
Those sub-volumes cropped from supervoxels of potential lesion areas still contain heterogeneous texture and rich contextual information, from which it is probable to learn effective feature representation for the target task of disease classification. In this study, we further propose a calibrated CPC scheme on the sub-volumes to embed the rich contextual information into 3D neural networks.

\begin{figure}[t]
  \begin{center}
    \includegraphics[width=0.75\columnwidth]{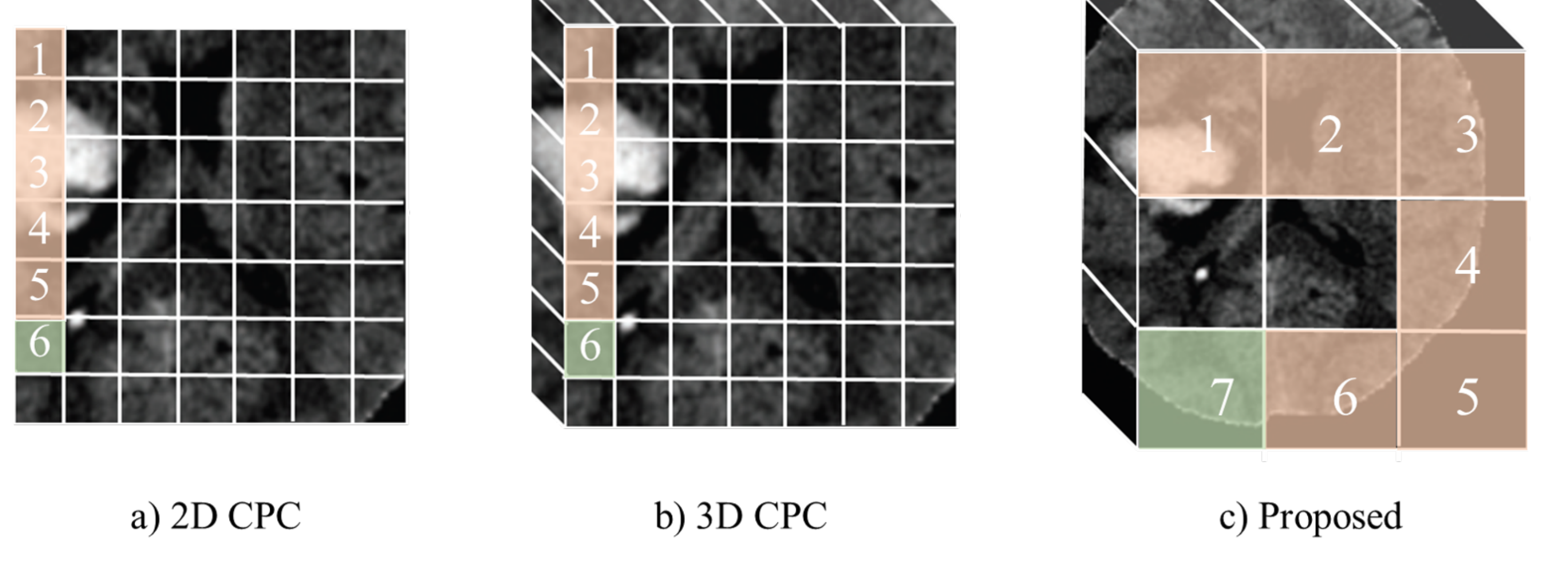}
    \caption{The sequence path for support generation adopted by our proposed method and conventional contrastive predictive coding methods. a) The path proposed in \cite{henaff2019dataefficient}. b) The extension of \cite{henaff2019dataefficient}. (c) The `U' shape path adopted in our TCPC. The cube sizes in a) and b) are relatively small, which contain limited information. The content of medical images has fewer variations (e.g, texture, grayscale), compared to natural images; hence, a relative larger cube size is more appropriate.}
    \label{proxy}
  \end{center}
\end{figure}

\subsubsection{Proxy task.} As shown in Fig.~\ref{framework}, the sub-volume is divided into $3 \times 3$ cubes as predictive units. Different from the typical CPC, which constitutes the support features along a straight path (Fig.~\ref{proxy}(a) and (b)), we propose a `U' shape path for the construction of support set; this way more contextual information of surrounding tissues can be included to better characterize the lesions. The difference between 2D CPC, 3D CPC (a simple extension of CPC from 2D to 3D) and our TCPC is illustrated in Fig.~\ref{proxy}, where the orange cubes are used as support set and the green one is for predictive coding. 

Assuming the feature extracted by a backbone network from the cube $i$ as $z_i$, the support set contained $N$ features can be represented as $Z=\{z_i,\,i=1,...,N\}$ and the ground-truth for the predictive coding is $z_{N+1}$. A recurrent neural network (RNN) $\psi$ is adopted to predict the feature vector ($z'_{N+1}$) of cube $N+1$ based on the support features, which can be formulated as:
\begin{equation}
  z'_{N+1}=\psi_{RNN}(Z)
\end{equation}

\subsubsection{Feature extractor.} A feature extractor is utilized to encode the features for support set and yields the corresponding ground-truth for the predictive coding. A sub-volume can be divided into $3\times 3$ cubes $x_{i,j}(i,j=1,2,3)$ with a 50\% overlap. The backbone network of feature extractor can be any 3D CNNs (e.g., 3D ResNet). Given the network function of feature extractor $f(.)$ and input cubes $x$, the features ($z$) of cubes can be generated via:
\begin{equation}
  z_{i}=f(x_{i})\quad i=1,...,N
\end{equation}

% We adopt a softmax layer to compute the probability assigned to the target, and use binary-cross-entropy loss to evaluate this probability.
\subsubsection{Contrastive loss.} The purpose of the proxy task is to recognize the target $z_{N+1}$ from other representations $z_i,\, i \in \{1,...,N\}$. Hence, a contrastive loss is adopted in our proxy task for network optimization, which can be defined as:
\begin{equation}
  Loss=-\sum_{i}^{z_{i}\neq z_{N+1}}
  \frac{z_{N+1}\log z'_{N+1}+(1-z_{N+1})log(1-z'_{N+1})}
  {z_{i}\log z'_{N+1}+(1-z_{i})log(1-z'_{N+1})}
\end{equation}
The $\{z_{i}|z_{i} \neq z_{N+1}\}$ denotes negative representations in the mini-batch. This loss is inspired by InfoNCE \cite{denoord2018representation} and proved to have the ability to maximize the mutual information.

\subsection{Adapting pre-trained network for target task}
Our TCPC proxy task can not only enforce 3D neural networks to exploit a robust feature representation from raw data, but also embed the clinical prior-knowledges, e.g., lesion position, into the neural networks. The pre-trained weights can be directly transferred to the target task, e.g., brain hemorrhage classification, with a small amount of annotated data.

\section{Experiment}

\subsection{Dataset}
\subsubsection{Brain hemorrhage dataset.}
The dataset contains 1,486 brain CT volumes, which are collected from a collaborative hospital. The CT volumes are used to diagnose the pathological cause of cerebral hemorrhage, which can be categorized into four classes: aneurysm, arteriovenous malformation, moyamoya disease and hypertension. The CT volumes are of a uniform size ($270 \times 230 \times 30$ voxels). The sub-volumes of $30 \times 150 \times 150$ voxels are cropped from the CT and partitioned to $3 \times 3$ $30 \times 75 \times 75$ cubes with 50\% overlapping for our TCPC proxy task. The network pre-trained on TCPC proxy task is then transferred to the target task (i.e., cerebral hemorrhage classification).

\subsubsection{Lung cancer dataset.}
LUNA16 dataset is a subset of LIDC-IDRI \cite{armato2011LIDC}, which is the largest publicly available dataset for pulmonary nodules detection. The dataset contains 888 low-dose lung CT volumes.
%which are with the same resolution---$512 \times 512$ transverse plane and $0.74\times 0.74$ $mm^{2}$ spacing. 
The CT scans with a slice thickness greater than 2.5 $mm$ are excluded. The potential lung nodules are already given and labeled by four experienced radiologists as benign (negative) or malignant (positive). The CT volumes have a uniform resolution ($512 \times 512$) with different slice number. The sub-volumes of $48 \times 48 \times 48$ voxels are cropped for the proxy task. Our goal here is to automatically classify these potential nodules into the benign and the malignant.

\subsection{Experimental settings}
The separation ratio of training and validation for brain hemorrhage classification and lung nodule classification are set to 80:20 and 90:10, respectively. The average classification accuracy (ACC) and area under curve (AUC) are employed as metric for performance evaluation. For comparison, several baseline models are evaluated, including SSL approaches (3D Jigsaw puzzles~\cite{noroozi2016jigsaw_puzzles} and Rubik's cube~\cite{Zhuang_2019_MICCAI}) and transfer learning (TL) method (fine-tuning the model pre-trained using UCF101 with label annotations~\cite{soomro2012ucf101}).

% We first normalize the images into [0, 255], segment the input image into several pieces and then select the area with the highest mean intensity as center (large probably related with lesion area according to prior knowledge). For SLIC cropping, 

\subsubsection{Implementation details.}
We first adopt supervoxel to center potential lesion areas, which is generated by clustering voxels with similar intensity. We use SLIC for voxel clustering, where the segment number, the maximum number of k-means iterations, and the Gaussian smoothing parameter are set to $10\sim 20$, 10, and 10, respectively. The widely used 3D VGG \cite{Simonyan15_VGG} and 3D ResNet-18 \cite{HeK2016} are adopted as backbone networks in the experiments, which are consistent with the settings of existing self-supervised learning approaches \cite{noroozi2016jigsaw_puzzles,Zhuang_2019_MICCAI}. Due to the different input sizes of pre-training and fine-tuning, we use an adaptive average pooling layer by the end of backbone network to unify the output shape.

The proposed TCPC is implemented using PyTorch. The network is trained with a mini-batch size of 2. The initial learning rate for the proxy task and target task are set to $2e^{-5}$ and $15e^{-6}$, respectively. The Adam solver \cite{kingma2014adam} is used as the optimizer for network training. The baseline approaches involved in our study adopt the same training protocol for fair comparison.

% For lung nodule classification, we adopt rotation and transformation as data augmentation.

% \subsection{Performance of supervoxel cropping}
% For proxy task, we first adopt supervoxel for center area (large probably related with lesion area) location. We segment the input image into 10 to 20 pieces and select the segmentation label with the highest mean intensity as center area. The results are shown in Fig.~\ref{super}. The results demonstrate that the unsupervised over-segmentation method is capable of sketchily localizing the lesion area. Such inclusion enables the features learned in the proxy task better transferred to the target task.
% \begin{figure}
%   \begin{center}
%     \includegraphics[width=4.5in]{figure/super.pdf}
%     \caption{Samples of center area location. The left picture of each pair is one slice of source 3D images and the right are the obtaining area and cropping boxes. We randomly crop 3D patches related with those central part.}
%     \label{super}
%   \end{center}
% \end{figure}

\subsection{Performance of proxy task}
\begin{figure}[!tb]
\begin{center}
\includegraphics[width=0.75\columnwidth]{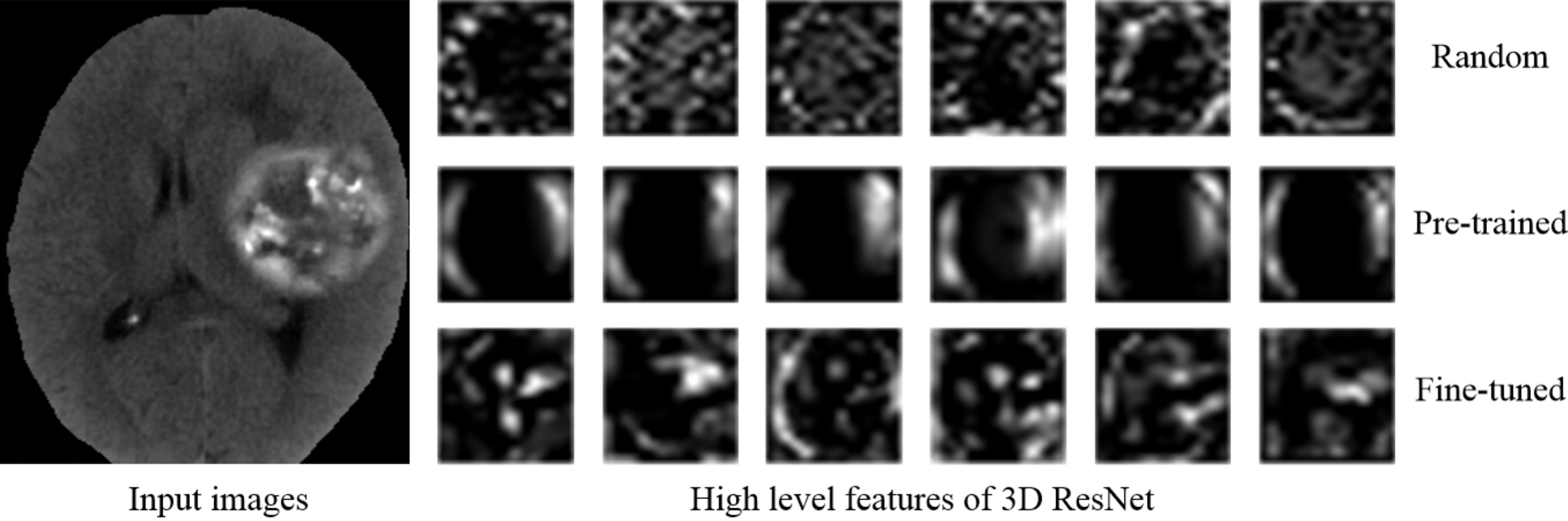}
\caption{Visualization of feature maps generated by the train-from-scratch model (`random'), the pre-trained model of TCPC and fine-tuned model. Rows 1, 2 and 3 represent six top-energy response channels generated by three different models, respectively.}
\label{feature_visual}
\end{center}
\end{figure}

For better evaluating the performance of our proxy task, the feature maps yield by the deep layers of 3D ResNet are presented in Fig.~\ref{feature_visual}. The results show that the pre-trained model (i.e., TCPC) captures the contextual information and task-related information. In particular, as shown in Fig.~\ref{feature_visual}, the lesion area of the input image is the brighter area on the right. The neural network is observed to pay attention to the brain boundary and lesion areas (the second row of Fig.~\ref{feature_visual}) after TCPC pre-training. In contrast, the features yield by a randomly initialized network contain less semantic information. Furthermore, the fine-tuned network generates the feature maps (the third row of Fig.~\ref{feature_visual}) with similar activation patterns to the one learned from TCPC proxy task, which demonstrates the benefit provided by our TCPC to the target task.

% ----------------
% figure feature_visual
% ----------------

\subsection{Fine-tuning network on target tasks}

\subsubsection{Brain hemorrhage.}
We fine-tune the pre-trained networks on the task of cerebral hemorrhage classification. The average classification accuracy (ACC) and area under curve (AUC) of different backbone networks trained with TCPC and other baseline approaches are presented in Table~\ref{hemorrhage_luna}. It can be observed that our TCPC can significantly boost the classification accuracy of backbone models. The TCPC pre-trained 3D VGG and ResNet-18 achieve ACCs of 78.38\% and 88.17\%, respectively, which are 6.08\% and 7.09\% higher than the train-from-scratch model. Similar improvements are observed in terms of AUC---the highest AUCs of 0.921 and 0.955 are obtained by our TCPC pre-trained 3D VGG and ResNet-18, respectively. The approaches that learn structural information of organs, e.g., 3D CPC, 3D Jigsaw puzzles, and Rubik's cube, yield lower improvements to classification accuracy, compared to our TCPC.

% ----------------
% Table classification results
% ----------------
\begin{table}[!tb]
\caption{The results of cerebral hemorrhage classification by using 100\% labeled data in fine-tuning and lung nodule classification using 1\% labeled data in fine-tuning. The performances of TCPC and other different approaches are evaluated by the average classification accuracy (ACC\%) and the area under curve (AUC).}
\label{hemorrhage_luna}
\centering
\begin{tabular}{l|cccc|cccc}
    \hline
    Dataset   & \multicolumn{4}{c|}{Brain Hemorrhage}    & \multicolumn{4}{c}{Lung Nodule}  \\
    \hline
    Model   & \multicolumn{2}{c|}{3D VGG}    & \multicolumn{2}{c|}{3D ResNet-18} & \multicolumn{2}{c|}{3D VGG}    & \multicolumn{2}{c}{3D ResNet-18}   \\ 
    \cline{2-9}
      &    ACC    &    AUC     &    ACC    &    AUC  &    ACC    &    AUC     &    ACC    &    AUC  \\
    \hline\hline
    Train-from-scratch & 72.30 & .889    & 81.08 &	.937  & 84.06 & .911  &  94.18 &	.982    \\
    UCF101 pre-trained~\cite{soomro2012ucf101} & 74.66 & .890    & 80.07 & .910  & 95.07 & .986  &  97.57 & .993     \\
    3D CPC & 76.35 &	.895  & 83.79  &  .921   & 92.19 &	.966  & 95.70  &  .989      \\
    3D Jigsaw puzzles \cite{noroozi2016jigsaw_puzzles}  & 73.31 & .891    & 85.81  & .951  & 89.43 & .940    & 98.12  & .992     \\
    Rubik's cube \cite{Zhuang_2019_MICCAI}      & 77.36 & .902    & 87.50  & .945   & 86.42 & .924    & 97.33  & .993     \\
    Ours TCPC  & 78.38 & \bf .921 &  88.17 & \bf .955  & 95.86 & .987 & 98.88 & \bf .996   \\
    \hline
   UCF101 + TCPC & \bf 79.39 & .920 & \bf 89.19 & \bf	.955 & \bf 97.85  & \bf .993  & \bf 99.40 & .995  \\\hline
\end{tabular}
\end{table}

\subsubsection{Lung cancer.}
To further demonstrate the effectiveness of our proxy task, we fine-tune the TCPC pre-trained networks on the lung cancer dataset. Since the original dataset is large, which neutralizes the improvement generated by pre-training, we constitute a subset (1\%) from the LUNA 16 as the fine-tuning training set for this experiment. The backbone is pre-trained on the whole training data without annotations and fine-tuned on the 1\% labeled data. The experimental results are shown in Table~\ref{hemorrhage_luna}. Our TCPC pre-trained 3D VGG and 3D ResNet-18 achieve ACCs of 95.86\% and 98.88\%, respectively, which outperform the listed baselines. The 3D Jigsaw and Rubik's cube are observed to yield relatively small improvements to backbone on this dataset. The underlying reason may be the small cube size of lung nodule, which leads to limited information contained in their partitioned tiles and cubes.

\subsubsection{UCF101 + TCPC.}
It can be observed from Table~\ref{hemorrhage_luna} that the UCF pre-trained model yields good performance, especially on lung nodule dataset. This suggests that though there exists gap between medical and natural images, transfer learning from a large natural image dataset is beneficial too. To further narrow down the gap and boost classification accuracy, we conduct an experiment that combines TL and SSL into a cascade. The 3D network is first initialized using the weights of the network trained for the UCF101 task, then pre-trained on the TCPC proxy task, and finally fine-tuned on the target task. The proposed UCF101 + TCPC yields an average 1\% improvement. Specifically, the proposed UCF101 + TCPC increases the ACCs of 3D VGG and ResNet-18 to 97.85\% and 99.40\% on lung nodule dataset, respectively. To the best of our knowledge, this marks \textbf{the first attempt} in the literature to explore the integration of TL and SSL methods for medical image analysis. In the future, we plan to explore further along this line in order to better address the deficient annotation issue.
%  The underlying reason may be the samples of lung cancer dataset are sub-volumes cropped from the whole lung, which result in the diverse appearances, compared to the registered brain CT volumes. The diverse contents may narrow down the gap between medical and natural images; thereby, the feature representation learned from UCF101 benefits the classification of lung nodule. 

% ----------------
% figure results
% ----------------
\begin{figure}[!htb]
\begin{center}
\includegraphics[width=0.98\columnwidth]{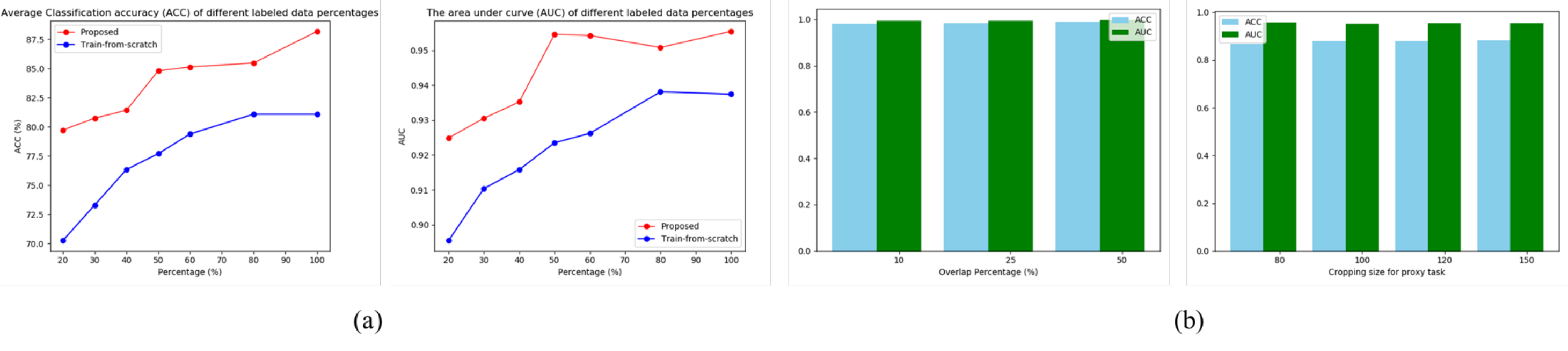}
\caption{(a) is the line chart of our TCPC and train-from-scratch network evaluated on brain hemorrhage dataset. Two indicators, i.e., average classification accuracy (ACC) (left figure) and area under curve (AUC) (right figure) are utilized for evaluation. The red curve and blue curve represent TCPC network and train-from scratch network, respectively. (b) illustrates the classification results by using different parameters. The left and the right figure of (b) are conducted by ResNet18 on 100\% brain hemorrhage dataset with different overlap percentages and different sub-volume sizes, respectively.}
\label{results1}
\end{center}
\end{figure}

\subsubsection{Performance of small training samples.}
In this section, we evaluate the performance of our TCPC with small training sets. Different amounts of original training set are extracted for pre-training and fine-tuning. The line chart of average classification accuracy (ACC) and area under curve (AUC) of TCPC pre-trained models with different amounts of fine-tuning data are presented in Fig.~\ref{results1}. The train-from-scratch model is also presented for comparison. The curves illustrate that our TCPC can consistently improve the ACC and AUC of backbone model, compared to train-from-scratch, especially when the training dataset is extremely small.
% It can be observed that the line of our TCPC tends to grow smoothly. Especially, the ACC and AUC improve significantly compared to train-from-scratch. 

\subsubsection{Evaluation of different parameters.}
The TCPC proxy task has two hyper-parameters, i.e., overlap ratio of cubes and cube size. The TCPC with different settings of these two hyper-parameters is evaluated in Fig.~\ref{results1}. Our TCPC is observed to be insensitive to the hyper-parameters. The larger cube size, which means that more contextual information is used for predictive coding, can increase the performance of our TCPC.
% We evaluate the performance of different images overlap in cells generation. Fig.~\ref{parameters}(a) shows that our TCPC has little insensitivity of parameters variation. We note that 50\% is the most common setting in traditional CPC. The larger cell size of our proposal may decrease the influence. In addition, we evaluate the effect of different cropping size for our proxy task, which shown in Fig.~\ref{parameters}(b). We discover that our method still holds a high performance with a range of cropping sizes. Such results offer a foundation that our method has little sensitivity with respect to cropping size.

\section{Conclusion}
In this paper, we develop a novel task-related self-supervised learning method for 3D medical images. By adopting a novel proxy task based on SLIC and CPC, the unsupervised learning features are intensive and more related to target task. We evaluate our proposed method on private and public 3D medical image datasets for classification tasks, and the results demonstrate efficiency and powerfulness for performance boost. Experiments empirically prove that task-related SSL approaches are effective tools in medical image analysis.
%We also discover the benefits of the combination of TL and SSL, and the exploration of such strategy will be our future work.


\bibliographystyle{splncs04}

\end{document}